\documentclass{article}
\usepackage[preprint]{spconf}
\usepackage{amsmath,graphicx}

\usepackage{caption}
\usepackage{subcaption}
\usepackage{adjustbox}
\usepackage[normalem]{ulem}
\useunder{\uline}{\ul}{}
\usepackage{multirow}
\usepackage{amssymb}
\usepackage{amsmath}
\usepackage{cleveref}

\copyrightnotice{\begin{minipage}[b]{2.05\linewidth}~~\\~~\\\copyright 2022 IEEE. Personal use of this material is permitted. Permission from IEEE must be obtained for all other uses, in any current or future media, including reprinting/republishing this material for advertising or promotional purposes, creating new collective works, for resale or redistribution to servers or lists, or reuse of any copyrighted component of this work in other works\end{minipage}}

\title{Bina-Rep Event Frames: a Simple and Effective Representation for Event-based cameras}
\name{Sami Barchid$^{\star}$ \qquad José Mennesson$^{\dagger, \star}$ \qquad Chaabane Djéraba$^{\star}$}
  
  \address{
    $^{\star}$ Univ. Lille, CNRS, Centrale Lille, UMR 9189 CRIStAL \\
    $^{\star}$ F-59000 Lille, France \\
    $^{\dagger}$ IMT Nord Europe, Institut Mines-Télécom, Centre for Digital Systems \\
    $^{\dagger}$   F-59000 Lille, France \\
    \{sami.barchid, chabane.djeraba\}@univ-lille.fr,  jose.mennesson@imt-nord-europe.fr
    }

\begin{document}
%
\maketitle
\begin{abstract}
This paper presents “Bina-Rep”, a simple representation method that converts asynchronous streams of events from event cameras to a sequence of sparse and expressive event frames. By representing multiple binary event images as a single frame of $N$-bit numbers, our method is able to obtain sparser and more expressive event frames thanks to the retained information about event orders in the original stream. Coupled with our proposed model based on a convolutional neural network, the reported results achieve state-of-the-art performance and repeatedly outperforms other common event representation methods. Our approach also shows competitive robustness against common image corruptions, compared to other representation techniques.
\end{abstract}
\begin{keywords}
Event Cameras, Neuromorphic Computer Vision, Convolutional Neural Networks, Object Recognition
\end{keywords}

\section{Introduction}
\label{sec:introduction}
Event-based cameras are neuromorphic sensors that produce asynchronous streams of events from brightness changes, which efficiently encodes temporal information. These sensors draw a lot of interest thanks to their great advantages such as low latency, sparse representation, low energy consumption, etc \cite{survey_dvs}. An effective and popular strategy is to apply conventional vision algorithms by converting a sequence of events into an event image accumulated during a certain time interval \cite{binary_event_image,event_histogram}. However, common event representation methods are either limited to inexpressive binary frames \cite{binary_event_image}, or event counts histograms \cite{event_histogram} that do not express any order in the accumulated sequence of events. In this paper, we propose "Bina-Rep" (for "Binary-Representation"), a simple approach to convert events to a sequence of event frames that are sparser and more expressive than other popular methods.

Our main contributions are as follows: \textit{(i)} Bina-Rep event frames, a new method to effectively apply conventional vision algorithms in event cameras, such as the proposed simple model based on Convolutional Neural Network (CNN); \textit{(ii)} a comparative study (against other representation techniques or state-of-the-art) reports competitive results in favor of our Bina-Rep method on both clean and corrupted data.
\section{Related Works}
\label{sec:related_works}

Due to the asynchronous nature of event streams produced by event-based sensors, several methods exist to transform them into alternative representations. In this section, we briefly review some of the most important strategies for event representations. 

\textbf{Individual Events} directly use the asynchronous events. Such methods rely on asynchronous processing algorithms such as Spiking Neural Networks \cite{plif}. This approach is highly efficient and perfectly integrates with event cameras, but is still limited in terms of results compared to traditional frame-based vision \cite{sami_decolle}.

\textbf{Time Surfaces} \cite{lagorce2016hots, ncars, dist} aim to extract the temporal information contained in a stream of events and explicitly expose it in 2D surfaces or images. Such representation tends to be less sensitive to noisy events and can be updated asynchronously \cite{survey_dvs}.

\textbf{Event Frames} \cite{binary_event_image, event_histogram} are simple representations that accumulate events into images to feed conventional vision algorithms (e.g. CNNs \cite{resnet}). Although no temporal information is exposed \cite{survey_dvs}, it is a popular method due to the intuitive representation of edge maps. In addition, the absence of explicit temporal information does not prevent them from achieving good results, even in truly neuromorphic datasets \cite{jason}.

Our Bina-Rep method can be categorized as following the \textbf{Event Frames} strategy. For the rest of the paper, we mainly compare our method with other event frame representations.
\section{Bina-Rep Event Frames}
\label{sec:binarep}
In this section, we define Bina-Rep, our representation method based on event frames. In addition, we highlight the differences with similar representation methods from previous works \cite{event_histogram, binary_event_image}. Finally, we introduce a simple model for event-based classification using sequences of 2D event representations.

\subsection{Definition}
\label{subsec:method_overview}
The output of an event camera of resolution $H \times W$ can be formulated as a sequence of events $\mathcal{E} = \{ e_i \}$, where $e_i = (x_i, y_i, t_i, p_i)$ represents a brightness change of polarity $p_i \in \{-1, 1\}$ occurring at time $t_i$ and pixel location $(x_i, y_i)$. A well-known representation method \cite{binary_event_image} aims to obtain a series of $T$ binary event frames $S \in \mathbb{R}^{C \times H \times W}$, where $C = 2$ for On/Off brightness changes. To do so, the events are accumulated in a defined time interval. 

The main idea behind our approach is that combining $N$ binary frames produces a sequence of $N$ bits for each pixel. For each pixel, we propose to consider the sequence of $N$ bits as an $N$-bit representation in other number systems. For instance, 8 binary frames can be represented as a single 8-bit unsigned integer frame.

In addition, our method can have the same usage as binary event frames, by producing a sequence of $T$ frames of $N$-bit numbers. Indirectly, this makes a sparse representation that relies on $T \times N$ binary event images.

\subsection{Differences with Similar Representations}
\label{subsec:difference_with}

In some ways, our method shares strong similarities with binary event images \cite{binary_event_image}, and event histograms \cite{event_histogram}. Similar to binary event images, our approach informs about the presence (or absence) of events during the time interval. It also has a notion of the number of events occurring during this interval (similar to event histograms), since it represents the accumulation of $T$ binary event images.

On the other hand, the $N$-bit number representation can be highly influenced by the order of events on the $N$ original binary event images. Our approach induces a sparser representation (i.e. $T$ frames instead of $T \times N$, assuming the subsequent vision algorithm is not quantized to deal with binary data), and more expressivity since the order of events can drastically change the value of an $N$-bit number. Still, these values indirectly represent the number of events accumulated during the time interval, even if it saturates faster than event histograms.

\begin{figure}
     \centering
     \begin{subfigure}[b]{0.20\textwidth}
         \centering
         \includegraphics[width=\textwidth]{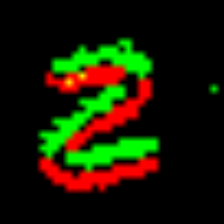}
         \caption{Binary Event Image \cite{binary_event_image}}
         \label{fig:frames_time}
     \end{subfigure}
     \hfill
     \begin{subfigure}[b]{0.20\textwidth}
         \centering
         \includegraphics[width=\textwidth]{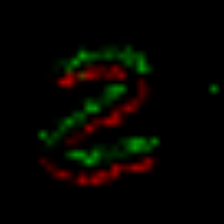}
         \caption{Bina-Rep event frame}
         \label{fig:bit_encoding}
     \end{subfigure}
     \caption{Comparative visualizations of a binary event image and a Bina-Rep event frame}
     \label{fig:comp_visu}
\end{figure}

To summarize, our proposed Bina-Rep event frames encode three information thanks to the $N$-bit number representation: \textit{(i)} the presence of events during a certain time interval, \textit{(ii)} the accumulated number of these events, and \textit{(iii)} the order of these events in the sequence. \Cref{fig:comp_visu} illustrates a comparative visualization between a Bina-Rep event frame and a binary event image. While the binary image is saturated and shows coarse edges, pixel values of a Bina-Rep event frame are more nuanced due to the encoded order of occurring events.

\subsection{Model for Event-based Recognition}
\label{subsec:model_for_eventbased_reco}
We formulate our method as follows: given an event-based input tensor $\mathbf{S} = \{ S\}^T$ composed of  $T$ 2D event representations, the objective is to predict the related label $c$ of the captured scene (e.g. a category of object, a given action performed by a person, \dots).

\begin{figure}
     \centering
     \includegraphics[width=0.48\textwidth]{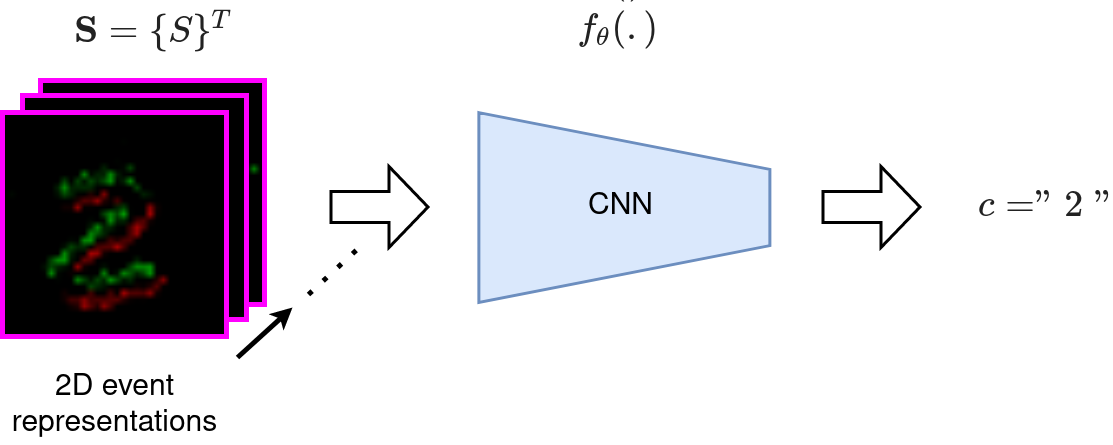}
     \caption{Overview of our model for event-based recognition}
     \label{fig:overview_model}
\end{figure}

To do so, a CNN architecture $f_{\theta}(\dot)$ is trained such that $c = f_{\theta}(\mathbf{S})$, where $\theta$ is the set of learnable parameters (weights, bias, etc). Since the architecture is a 2D CNN, the sequence of 2D event representations is concatenated along the channel axis. In this way, different representations can be easily compared depending on the performance achieved by the CNN. \Cref{fig:overview_model} illustrates an overview of this pipeline.

\section{Experiments}
\label{sec:experiments}

\subsection{Datasets and Metrics}
\label{subsec:datasets_and_metrics}
We evaluate Bina-Rep and other common event representation methods in several popular event-based recognition datasets: N-MNIST \cite{nmnist}, DVS-Gesture \cite{dvsgesture}, N-Cars \cite{ncars}, and CIFAR10-DVS \cite{cifar10_dvs}. \Cref{tab:dataset_summary} briefly describes these benchmarks. We report the top-1 accuracy (expressed in \%) to evaluate and compare all experiments.

\begin{table}
\begin{adjustbox}{max width=0.48\textwidth}
\begin{tabular}{c|ccc}
\textbf{Dataset}     & \begin{tabular}[c]{@{}c@{}}\# of\\ training samples\end{tabular} & \begin{tabular}[c]{@{}c@{}}\# of\\ validation samples\end{tabular} & \begin{tabular}[c]{@{}c@{}}\# of\\ classes\end{tabular} \\ \hline
\textbf{N-MNIST} \cite{nmnist}     & 60000                                                            & 10000                                                              & 10                                                      \\
\textbf{CIFAR10-DVS} \cite{cifar10_dvs} & 8000                                                             & 2000                                                               & 10                                                      \\
\textbf{N-Cars} \cite{ncars}     & 15422                                                            & 8607                                                               & 2                                                       \\
\textbf{DVS-Gesture} \cite{dvsgesture} & 1077                                                             & 264                                                                & 11                                                     
\end{tabular}
\end{adjustbox}
\caption{Summary of the event datasets used in our study}
\label{tab:dataset_summary}
\end{table}

\subsection{Implementation Details}
\label{subsec:implementation_details}
We run our experiments on PyTorch \cite{pytorch} using an NVIDIA Tesla P100 GPU. All models are trained using cross-entropy loss from random initialization (i.e. no pre-training) during 60 epochs. For each dataset, we resize the resolution of the camera to $224 \times 224$. We use an Adam \cite{adam} optimizer with an initial learning rate of $0.001$, and divide it by $10$ after every 15 epochs. We use a ResNet-18 \cite{resnet} as the CNN in our baseline model described in \Cref{subsec:model_for_eventbased_reco}. In \Cref{tab:repr_params}, we summarize the studied event representation methods as the input tensor that is fed to our CNN model (discussed in \Cref{subsec:method_overview}). As for Bina-Rep event frames, we use an 8-bit unsigned integer representation for all our experiments (i.e. $N = 8$).

\begin{table}
\centering
\begin{tabular}{c|ccc}
\textbf{\begin{tabular}[c]{@{}c@{}}Event Representation\end{tabular}} & \textbf{C}         & \textbf{T} & \textbf{\begin{tabular}[c]{@{}c@{}}\# \\ of channels\end{tabular}} \\ \hline
\textbf{Voxel Grid} \cite{voxel}                                                     & -                  & 10         & 10                                                                 \\
\textbf{Binary Event Images} \cite{binary_event_image}                                            & \multirow{4}{*}{2} & 10         & 20                                                                 \\
\textbf{Event Histogram} \cite{event_histogram}                                                &                    & 1          & 2                                                                  \\
\textbf{Bina-Rep} (ours)                                                      &                    & 1          & 2                                                                  \\
\textbf{Bina-Rep} (ours)	                                                       &                    & 3          & 6                                                                 
\end{tabular}
\caption{Parameters for each studied event representation. Aside from Voxel Grid \cite{voxel}, the $T$ event frames are concatenated with the $C=2$ on/off channels of event-based cameras to obtain the final number of channels for the input tensor of our CNN}
\label{tab:repr_params}
\end{table}	

\subsection{Performance Analysis}
\label{subsec:performance_analysis}
\subsubsection{Event Frames Methods Comparison}
We conduct a comparative study between Bina-Rep event frames and other popular representation methods that transform event-based sequences into images. Moreover, we investigate the performance of Bina-Rep event-frames in both single frame ($T=1$) and multiple frames ($T=3$) settings. Our results are reported in Table \ref{tab:representation_comparison}. In general, our proposed method achieves competitive results for all datasets. More precisely, it outperforms other methods on N-MNIST and shows the second-best performance on N-Cars and DVS-Gesture. For our approach, we observe better performance on the multiple frames settings ($T=3$) in general, which can be explained by a more precise decomposition of the original stream of events.

Contrary to other studied datasets, DVS-Gesture has been shown to be truly neuromorphic compared to the others \cite{jason} because it integrates temporal information from longer gesture sequences while other methods shortly move an event-based camera on static objects to obtain only spatial information. Consequently, the reported results on DVS-Gesture are important to investigate since it suggests the capacity of an event representation method to handle temporal information. The reported results vary importantly across the compared approaches, which shows a significant difference in the capacity to handle temporal information. Our Bina-Rep event frames with $(T = 3)$ achieves second-best accuracy, which highlights its efficiency to deal with temporal information. Our approach with $T=3$ also achieves better performance than Binary Event Images with $T=10$, confirming the better expressivity and sparser representation explained in Section \ref{subsec:difference_with}. Interestingly, Event Histogram outperforms other methods, suggesting that it can effectively deal with temporal information without specifically leveraging temporal information. On the other hand, our single frame setting achieves poor performance. It can be explained by the fact that a single Bina-Rep event frame cannot integrate a high quantity of events from longer temporal sequences because it would quickly saturate the $N$-bit representation of pixels.

\begin{table}
\begin{adjustbox}{max width=0.48\textwidth}
\begin{tabular}{c|cccc}
\textbf{Representation}      & \textbf{\begin{tabular}[c]{@{}c@{}}N-MNIST\\ \cite{nmnist}\end{tabular}} & \textbf{\begin{tabular}[c]{@{}c@{}}CIFAR10-DVS\\ \cite{cifar10_dvs}\end{tabular}} & \textbf{\begin{tabular}[c]{@{}c@{}}N-Cars\\  \cite{ncars}\end{tabular}} & \textbf{\begin{tabular}[c]{@{}c@{}}DVS-Gesture\\ \cite{dvsgesture}\end{tabular}} \\ \hline
\textbf{Voxel Grid }\cite{voxel}               & {\ul 99.43}                                                  & \textbf{94.00}                                                   & 90.71                                                       & 79.55                                                            \\
\textbf{Binary Event Images} \cite{binary_event_image} & 99.36                                                        & 93.45                                                            & \textbf{92.48}                                              & 82.95                                                            \\
\textbf{Event Histogram} \cite{event_histogram}     & 99.14                                                        & {\ul 93.65}                                                      & 91.77                                                       & \textbf{90.91}                                                   \\ \hline
\textbf{Bina-Rep $T=1$} (ours) & 99.34                                                        & 92.4                                                             & {\ul 92.04}                                                 & 80.68                                                            \\
\textbf{Bina-Rep $T=3$} (ours) & \textbf{99.52}                                               & 93.25                                                            & 90.74                                                       & {\ul 87.88}                                                     
\end{tabular}
\end{adjustbox}
\caption{Comparison of common event representation methods on event-based datasets}
\label{tab:representation_comparison}
\end{table}

\subsubsection{Comparison with State-of-the-art}
We compare the performance of our CNN trained on Bina-Rep event frames with results from state-of-the-art methods (shown in Table \ref{tab:sota_comp}). We observe that our method outperforms previous works in many datasets, showing the efficacy of our simple model to perform event-based recognition tasks. However, it is important to mention that the models used can differ importantly from one method to the other. For instance, DiST\cite{dist} uses a ResNet-34 CNN, while HATS \cite{ncars} is based on a linear SVM.

\begin{table}
\begin{adjustbox}{max width=0.48\textwidth}
\begin{tabular}{c|ccc}
\textbf{Methods}    & \textbf{\begin{tabular}[c]{@{}c@{}}N-MNIST\\\cite{nmnist}\end{tabular}} & \textbf{\begin{tabular}[c]{@{}c@{}} CIFAR10-DVS\\ \cite{cifar10_dvs}\end{tabular}} & \textbf{\begin{tabular}[c]{@{}c@{}}N-Cars \\ \cite{ncars}\end{tabular}} \\ \hline
HATS    \cite{ncars}            & 99.1                                                           & 52.4                                                               & 90.2                                                         \\
PLIF SNN     \cite{plif}       & \textbf{99.6}                                                  & 74.8                                                               & -                                                            \\
DiST        \cite{dist}        & -                                                              & 62.57                                                              & 90.80                                                        \\
EvS-B   \cite{graph}            & -                                                              & 68.0                                                               & \textbf{93.1}                                                \\ \hline
Bina-Rep $T=1$ (ours) & 99.34                                                          & {\ul 92.4}                                                         & {\ul 92.04}                                                  \\
Bina-Rep $T=3$ (ours) & {\ul 99.52}                                                    & \textbf{93.25}                                                     & 90.74                                                       
\end{tabular}
\end{adjustbox}
\caption{Comparison between our proposed method and the state-of-the-art methods on different datasets}
\label{tab:sota_comp}
\end{table}

\subsection{Robustness against Common Corruptions}
\label{subsec:robustness}
In addition to competitive results on clean datasets, it is also important to verify that an event representation method is not fragile to common altered scenarios. Therefore, we study the impact of simulating two common corruptions on the N-Cars \cite{ncars} validation set (after training on clear data). The two following corruptions are investigated: \textbf{Background Activity:} corruption that occurs due to thermal noise and junction leakage current in an event-based sensor. As a result, output events are produced without any light intensity changes \cite{ba}. It is important to mention that this corruption is already present in clear datasets. Therefore, this corruption is aimed at increasing the background activity noise. \textbf{Occlusion:} corruption that simulates the presence of an object occluding the scene. In this work, we simulate it by dropping events in an occlusion box placed at the center of the scene.

Inspired from \cite{benchmark_corruption}, we define 5 levels of severity for a specific corruption, where the level of severity increases the intensity of the corruption. For background activity, the level of severity represents the ratio of background events added to the original sequence. For occlusion, it represents the size of the occlusion box, in the percentage of the image resolution. Table \ref{tab:severity_params} describes the value of the parameters for each level of severity.

\begin{table}
\begin{tabular}{c|ccccc}
\multirow{2}{*}{\textbf{Corruption}} & \multicolumn{5}{c}{\textbf{Severity level}}                    \\
                                     & \textbf{1} & \textbf{2} & \textbf{3} & \textbf{4} & \textbf{5} \\ \hline
Background Activity                  & 0.5\%      & 0.8\%      & 1.0\%      & 2.0\%      & 3.0\%      \\
Occlusion                            & 35\%       & 45\%       & 50\%       & 60\%       & 70\%      
\end{tabular}
\caption{Parameter values of the studied corruptions for each severity level}
\label{tab:severity_params}
\end{table}

\begin{figure}
     \centering
     \begin{subfigure}[b]{0.35\textwidth}
         \centering
         \includegraphics[width=\textwidth]{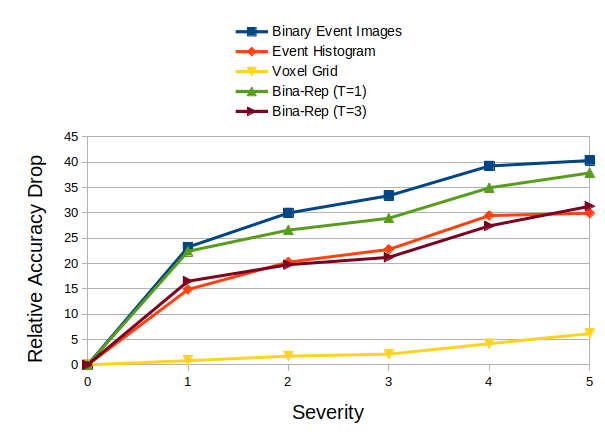}
         \caption{Background Activity noise}
         \label{fig:background_activity}
     \end{subfigure}
     \\
     \begin{subfigure}[b]{0.35\textwidth}
         \centering
         \includegraphics[width=\textwidth]{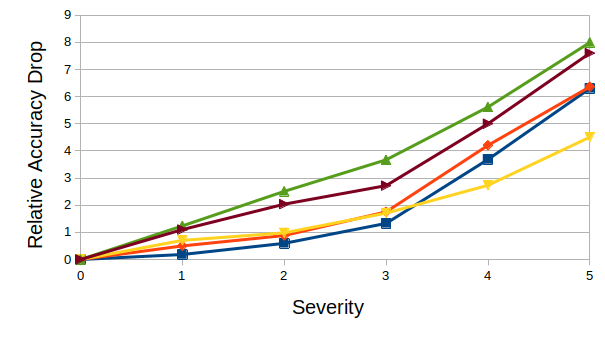}
         \caption{Occlusion corruption}
         \label{fig:occlusion}
     \end{subfigure}
     \caption{Relative Accuracy Drop scores for each event representation method}
     \label{fig:rad}
\end{figure}

For a given corruption, the Relative Accuracy Drop score is used to measure the sensitivity of each event representation method: $score = \frac{acc_{0} - acc_{i}}{acc_{0}} \times 100$, where $acc_i$ and $acc_0$ are the accuracy for the validation set corrupted with a severity level of $i$ and the accuracy on clean data, respectively. We report the Relative Accuracy Drop scores for each studied event representations in Figure \ref{fig:rad}. 

For Background Activity noise (shown in Figure \ref{fig:background_activity}), we observe the same evolution and similar scores for all methods aside from Voxel Grid \cite{voxel} that is greatly robust against this type of corruption. On the other hand, Bina-Rep shows better performance than Binary Event Frames on the $T=1$ settings and performs on par with Event Histogram \cite{event_histogram} with $T = 3$, which suggests higher robustness by increasing the number of Bina-Rep frames. It can be explained by the fact that a Bina-Rep frame expresses the order of arrival of events in addition to their occurrence (as explained in Section \ref{subsec:difference_with}). Therefore, a single Bina-Rep saturates because of background events, which harms the performance. By using multiple Bina-Rep frames, this influence is reduced and we report similar performance to Event Histogram. \cite{event_histogram}.

As for occlusion (shown in Figure \ref{fig:occlusion}), all methods seem to be quite robust to this type of corruption (the relative accuracy drop scores remain $< 9$ \%). The same evolution is observed as well. Voxel Grid still shows great robustness for high levels of sensitivity but follows the same pattern as other representations. Bina-Rep shows slightly worse scores but remains competitive compared to other event representation methods.
\section{Conclusion}
\label{sec:conclusion}
In this work, we proposed a new representation method as images for event cameras named "Bina-Rep". By simply representing multiple binary event frames as $N$-bit number systems, additional information on the original event sequence is encoded in a sparse representation. In addition, we presented a simple CNN-based model and investigated the performance of Bina-Rep and other common event representation approaches. Our model coupled with Bina-Rep event frames repeatedly outperforms both state-of-the-art methods and other event representation methods coupled with our CNN model. Finally, we study the sensitivity of the event representation methods to common image corruptions and find that Bina-Rep event frames show competitive robustness. We believe that our novel Bina-Rep approach can be an efficient option for sparser and more expressive event-based algorithms for neuromorphic vision.

\bibliographystyle{IEEEbib}
\bibliography{refs}

\begin{thebibliography}{10}

\bibitem{survey_dvs}
Guillermo Gallego, Tobi Delbr{\"u}ck, Garrick Orchard, Chiara Bartolozzi, Brian
  Taba, Andrea Censi, Stefan Leutenegger, Andrew~J Davison, J{\"o}rg Conradt,
  Kostas Daniilidis, et~al.,
\newblock ``Event-based vision: A survey,''
\newblock {\em IEEE transactions on pattern analysis and machine intelligence},
  vol. 44, no. 1, pp. 154--180, 2020.

\bibitem{binary_event_image}
J{\"u}rgen Kogler, Christoph Sulzbachner, and Wilfried Kubinger,
\newblock ``Bio-inspired stereo vision system with silicon retina imagers,''
\newblock in {\em International Conference on Computer Vision Systems}.
  Springer, 2009, pp. 174--183.

\bibitem{event_histogram}
Ana~I Maqueda, Antonio Loquercio, Guillermo Gallego, Narciso Garc{\'\i}a, and
  Davide Scaramuzza,
\newblock ``Event-based vision meets deep learning on steering prediction for
  self-driving cars,''
\newblock in {\em Proceedings of the IEEE Conference on Computer Vision and
  Pattern Recognition}, 2018, pp. 5419--5427.

\bibitem{plif}
Wei Fang, Zhaofei Yu, Yanqi Chen, Timoth{\'e}e Masquelier, Tiejun Huang, and
  Yonghong Tian,
\newblock ``Incorporating learnable membrane time constant to enhance learning
  of spiking neural networks,''
\newblock in {\em Proceedings of the IEEE/CVF International Conference on
  Computer Vision}, 2021, pp. 2661--2671.

\bibitem{sami_decolle}
Sami Barchid, Jos{\'e} Mennesson, and Chaabane Dj{\'e}raba,
\newblock ``Deep spiking convolutional neural network for single object
  localization based on deep continuous local learning,''
\newblock in {\em 2021 International Conference on Content-Based Multimedia
  Indexing (CBMI)}. IEEE, 2021, pp. 1--5.

\bibitem{lagorce2016hots}
Xavier Lagorce, Garrick Orchard, Francesco Galluppi, Bertram~E Shi, and Ryad~B
  Benosman,
\newblock ``Hots: a hierarchy of event-based time-surfaces for pattern
  recognition,''
\newblock {\em IEEE transactions on pattern analysis and machine intelligence},
  vol. 39, no. 7, pp. 1346--1359, 2016.

\bibitem{ncars}
Amos Sironi, Manuele Brambilla, Nicolas Bourdis, Xavier Lagorce, and Ryad
  Benosman,
\newblock ``Hats: Histograms of averaged time surfaces for robust event-based
  object classification,''
\newblock in {\em Proceedings of the IEEE Conference on Computer Vision and
  Pattern Recognition}, 2018, pp. 1731--1740.

\bibitem{dist}
Junho Kim, Jaehyeok Bae, Gangin Park, Dongsu Zhang, and Young~Min Kim,
\newblock ``N-imagenet: Towards robust, fine-grained object recognition with
  event cameras,''
\newblock in {\em Proceedings of the IEEE/CVF International Conference on
  Computer Vision}, 2021, pp. 2146--2156.

\bibitem{resnet}
Kaiming He, Xiangyu Zhang, Shaoqing Ren, and Jian Sun,
\newblock ``Deep residual learning for image recognition,''
\newblock in {\em Proceedings of the IEEE conference on computer vision and
  pattern recognition}, 2016, pp. 770--778.

\bibitem{jason}
Laxmi~R Iyer, Yansong Chua, and Haizhou Li,
\newblock ``Is neuromorphic mnist neuromorphic? analyzing the discriminative
  power of neuromorphic datasets in the time domain,''
\newblock {\em Frontiers in neuroscience}, vol. 15, pp. 297, 2021.

\bibitem{nmnist}
Garrick Orchard, Ajinkya Jayawant, Gregory~K Cohen, and Nitish Thakor,
\newblock ``Converting static image datasets to spiking neuromorphic datasets
  using saccades,''
\newblock {\em Frontiers in neuroscience}, vol. 9, pp. 437, 2015.

\bibitem{dvsgesture}
Arnon Amir, Brian Taba, David Berg, Timothy Melano, Jeffrey McKinstry, Carmelo
  Di~Nolfo, Tapan Nayak, Alexander Andreopoulos, Guillaume Garreau, Marcela
  Mendoza, et~al.,
\newblock ``A low power, fully event-based gesture recognition system,''
\newblock in {\em Proceedings of the IEEE conference on computer vision and
  pattern recognition}, 2017, pp. 7243--7252.

\bibitem{cifar10_dvs}
Hongmin Li, Hanchao Liu, Xiangyang Ji, Guoqi Li, and Luping Shi,
\newblock ``Cifar10-dvs: an event-stream dataset for object classification,''
\newblock {\em Frontiers in neuroscience}, vol. 11, pp. 309, 2017.

\bibitem{pytorch}
Adam Paszke, Sam Gross, Soumith Chintala, Gregory Chanan, Edward Yang, Zachary
  DeVito, Zeming Lin, Alban Desmaison, Luca Antiga, and Adam Lerer,
\newblock ``Automatic differentiation in pytorch,''
\newblock in {\em NIPS-W}, 2017.

\bibitem{adam}
Diederik~P Kingma and Jimmy Ba,
\newblock ``Adam: A method for stochastic optimization,''
\newblock {\em arXiv preprint arXiv:1412.6980}, 2014.

\bibitem{voxel}
Alex~Zihao Zhu, Liangzhe Yuan, Kenneth Chaney, and Kostas Daniilidis,
\newblock ``Unsupervised event-based learning of optical flow, depth, and
  egomotion,''
\newblock in {\em Proceedings of the IEEE/CVF Conference on Computer Vision and
  Pattern Recognition}, 2019, pp. 989--997.

\bibitem{graph}
Yijin Li, Han Zhou, Bangbang Yang, Ye~Zhang, Zhaopeng Cui, Hujun Bao, and
  Guofeng Zhang,
\newblock ``Graph-based asynchronous event processing for rapid object
  recognition,''
\newblock in {\em Proceedings of the IEEE/CVF International Conference on
  Computer Vision}, 2021, pp. 934--943.

\bibitem{ba}
Yang Feng, Hengyi Lv, Hailong Liu, Yisa Zhang, Yuyao Xiao, and Chengshan Han,
\newblock ``Event density based denoising method for dynamic vision sensor,''
\newblock {\em Applied Sciences}, vol. 10, no. 6, 2020.

\bibitem{benchmark_corruption}
Dan Hendrycks and Thomas Dietterich,
\newblock ``Benchmarking neural network robustness to common corruptions and
  perturbations,''
\newblock {\em arXiv preprint arXiv:1903.12261}, 2019.

\end{thebibliography}

\end{document}